\documentclass[a4paper,twoside]{article}

\usepackage{graphicx}
\usepackage{float}
\usepackage{multirow}
\usepackage{amsmath,amssymb} 
\usepackage{hyperref}
\usepackage{graphicx}
\usepackage{amsmath}
\usepackage{subcaption}
\usepackage{color,soul}
\usepackage{multirow}
\usepackage{rotating}

\usepackage[width=122mm,left=12mm,paperwidth=146mm,height=193mm,top=12mm,paperheight=217mm]{geometry}

\usepackage{epsfig}
\usepackage{calc}
\usepackage{amssymb}
\usepackage{amstext}
\usepackage{amsmath}
\usepackage{amsthm}
\usepackage{multicol}
\usepackage{pslatex}
\usepackage{apalike}
\usepackage{SCITEPRESS}     

\usepackage{scalerel}

\begin{document}

\title{Exploring Deep Spiking Neural Networks \\ for Automated Driving Applications}

\author{\authorname{Sambit Mohapatra\sup{1}, Heinrich Gotzig\sup{1}, Senthil Yogamani\sup{2},  Stefan Milz\sup{3} and Raoul Z\"{o}llner\sup{4}}
\affiliation{\sup{1}Valeo Bietigheim, Germany}
\affiliation{\sup{2}Valeo Vision Systems, Ireland}
\affiliation{\sup{3}Valeo Kronach, Germany}
\affiliation{\sup{4}Heilbronn University, Germany}
\email{\{sambit.mohapatra,heinrich.gotzig,senthil.yogamani,stefan.milz\}@valeo.com, raoul.zoellner@hs-heilbronn.de}
}


\abstract{Neural networks have become the standard model for various computer vision tasks in automated driving including semantic segmentation, moving object detection, depth estimation, visual odometry, etc. The main flavors of neural networks which are used commonly are convolutional (CNN) and recurrent (RNN). In spite of rapid progress in embedded processors, power consumption and cost is still a bottleneck. Spiking Neural Networks (SNNs) are gradually progressing to achieve low-power event-driven hardware architecture which has a potential for high efficiency. In this paper, we explore the role of deep spiking neural networks (SNN) for automated driving applications. We provide an overview of progress on SNN and argue how it can be a good fit for automated driving applications. 
}

\onecolumn \maketitle \normalsize \vfill

\section{Introduction} \label{intro}

\textbf{Autonomous driving} is a rapidly progressing area of automobile engineering that aims to gradually reduce human interaction in automobile driving. Divided into 5 levels of autonomy, level 4 and 5 target the ultimate goal of automated driving, namely complete removal of human interaction in vehicle driving. The overall task of autonomous driving may be sub-divided into 3 key groups of activities - (1) Environmental sensing, (2) Environmental perception from sensor data and (3) Actuation of drive action according to perception. More often than not, it has been seen that the type of sensor and its output, define the approach most suitable for perception of the environmental from the sensor data.

CNN (Convolutional Neural Networks) has made huge leaps in accuracy for various computer vision tasks like object recognition and semantic segmentation \cite{siam2017deep}. They are also becoming dominant in geometric tasks like depth estimation, motion estimation, visual odometry, etc. It has played a major role in achieving high accuracy for various computer vision tasks which is critical for safe automated driving systems. However, they are computationally expensive and power consumption is becoming a bottleneck. For example, the recently announced Nvidia platform Xavier provides 30 Tera-ops (TOPS) of compute power but consumes 30 Watts. This necessitates an active cooling system which will consume more power and add to operating costs. 

SNN (Spiking Neural networks) have been progressing gradually as a power efficient neural network. The functional capabilities of SNN neuron model is discussed in detail in  \cite{chou2018algorithmic}. SNNs were proven to be effective in several problems but it remained less competitive compared to CNNs. Recently \cite{sengupta2018going} demonstrated that a deep SNN can achieve better accuracy than CNN on a challenging dataset ImageNet. A detailed overview of deep learning in SNN is discussed in \cite{tavanaei2018deep}.  \cite{wunderlich2018demonstrating} discuss the power consumption advantages of SNN when implemented in neuromorphic hardware. In \cite{zhou2018object}, SNN was shown to be effective for LIDAR object detection directly on analog signals. Motivated by the recent progress in SNN, we study the potential of SNN for automated driving applications in this paper. 

The rest of the paper is structured as follows. Section \ref{related} provides an overview of Spiking neural networks (SNN) and compares it with popular version of NNs namely CNNs and RNNs. Section \ref{snn4ad} discusses opportunities of SNNs for automated driving applications and provides motivating use cases. Finally, section \ref{conclusion} summarizes the paper and provides potential future directions.


\section{Related work on SNN} \label{related}
Sensors such as camera, lidar and radar generate enormous amounts of data.
Machine learning has proven to be highly successful in tasks involving such high dimensional data.
Since most objects in the environment can be grouped into certain classes such as pedestrian, cars etc, the data presents a pattern, which can be used to train classifiers that can then classify objects with great accuracy.
Historically, Convolutional Neural Networks (CNNs) have been the mainstay of all major machine learning approaches to object detection.
Some prominent models that make use of CNNs for object detection are Fast R-CNN \cite{girshick2015fast} which uses the fine-tuned CNN to extract features from object proposals and Support Vector Machines (SVM) to classify them.

R-CNN based algorithms use a two-step process for object detection namely - region proposal and region classification.
Recently one-shot methods have also been proposed such as YOLO \cite{redmon2016you} and SSD.
All these methods generate feature maps which make up the bulk of the computation and then classification of the feature maps.

Spiking Neural Networks (SNNs) are the most recent addition to the family of neural networks and machine learning. Considering the fact that they are still in a preliminary stage of research, large scale practical applications and implementations on hardware are rather few.
Some of the most notable applications include \cite{diehl2015fast} that use conversion techniques for converting a CNN into SNN to achieve impressive error rate of 0.9\% in MNIST digit recognition application.
Another notable application is \cite{hunsberger2015spiking} where Leaky Integrate and Fire (LIF) neuron model with smoothened response is used to convert a CNN to SNN for object recognition application on CIFAR-10 dataset.

\subsection{Overview of SNN }
\begin{figure}
    \centering
    \includegraphics[width=0.5\textwidth]{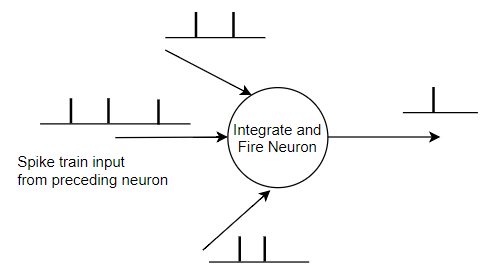}
    \caption{Spiking neuron model} 
    \label{fig:spikingneuron}
\end{figure}

Widely considered as the third generation of neural models, Spiking Neural models differ from conventional neural models in the very way that information is represented and processed by them.
This is inspired by information representation and processing in biological neurons where information is converted into a voltage spike train generally of equal amplitude. The duration and timing of the spikes encodes the actual information. In its very basic form, information arrives at a neuron from preceding neurons in the form of spikes, which are integrated over time. Once accumulated voltage reaches a certain threshold, a voltage spike is sent out as the output from the neuron. Figure \ref{fig:spikingneuron} illustrates a basic representation of a spiking neuron model. 

Spiking networks are capable of processing a large pool of data using a small number of spikes \cite{thorpe2001spike}.
Previous work has demonstrated that SNNs can be applied to all common tasks to which CNNs are applied and can do so in an effective way \cite{maass1997networks}.
Spiking neuron models are highly motivated by biological neurons and the way they function \cite{bois1848untersuchungen}.
There are 3 main characteristics:
(1) It accept inputs from many incoming synapses and produce single output spike.
(2) Inputs can be excitatory - if they increase the firing rate of a neuron or inhibitory - if they reduce firing rate of a neuron.
(3) The neuron model is governed by at least one state variable. In spiking models, the spike timings carry the information rather than the amplitude or shape \cite{gerstner2002spiking}.

A spike train can be described as 

\begin{equation}
    S(t)=\sum_{\scaleto{f}{4pt}}\delta(t-t_{\scaleto{f}{4pt}}) 
\end{equation}

where f = 1, 2, ... is the label of the spike 
$\delta$(.) is a Dirac function as defined below whose area is 1. 

\begin{equation}
\delta(t) =
    \begin{cases}
            1, &         \text{if } t=0,\\
            0, &         \text{if } t\neq 0.
    \end{cases}
\end{equation}
Neuron models are used to represent the dynamics of signal processing in a neuron mathematically.
In case of Spiking neurons, three commonly used models are - Hodgkin-Huxley model, Izhikevich model and Integrate and Fire model.
While the Hodgkin-Huxley model provides the closest modelling of actual biological neurons, it's mathematical complexity makes it unsuitable for use in applications.
A version of the Integrate and Fire model known as the Leaky Integrate and Fire (LIF) model is the most widely used neuronal model for spiking neurons as it provides a balance between mathematically complexity of implementation and closeness to biological process
\cite{gerstner2002spiking}.
LIFs are mathematically represented as:

\begin{equation}
C\frac{du}{dt}(t) = - \frac{u(t)}{R} + (i_{o}(t) + \sum w_j i_j(t) )
\end{equation}

where u(t): state variable (membrane potential),
C: membrane capacitance, 
R: input resistance, 
$i_{o}(t)$: is the external current, 
$i_{j}(t)$: is the input current from the j-th synaptic input
$w_j$: strength of the j-th synapse.
A neuron fires a spike at time t , if membrane potential u reaches threshold(v). Immediately after a spike the membrane potential is reset to a value less than the threshold and held for the time known as the refractory period.
SNN can be represented as a directed graph (V, S), with V being a set of neurons and S representing a set of syn¬apses \cite{maass1997networks}. The set V contains a subset of input and output neurons.\\

\noindent \textbf{Spiking network topologies:} \\

1. \textbf{Feedforward} networks - The data flows from input to output in a unidirectional manner across several layers.
Applications include sensory systems, e.g. in vision \cite{escobar2009action}, olfaction \cite{fu2007pattern} or tactile sensing \cite{cassidy2006biologically}.

2. \textbf{Recurrent} networks - In this case, neuron groups have feedback connections. This allows dynamic temporal behavior of the network. However, this feedback arrangement makes control more difficult in such networks \cite{hertz1991introduction}.

3. \textbf{Hybrid} networks - Some of the neurons have feedback connections while other are connected in a feed-forward fashion. \\

\noindent \textbf{Spike coding techniques:} \\

\begin{figure}[!t]
\centering
\begin{subfigure}{0.5\textwidth}
    \centering
    \includegraphics[scale=0.8]{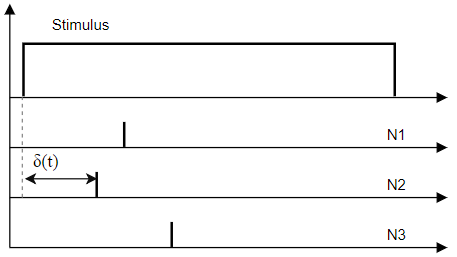}
    \caption{\textcolor{black}{Time to first spike \newline}}
\end{subfigure}%
\qquad
\begin{subfigure}{0.5\textwidth}
    \centering
    \includegraphics[scale=0.8]{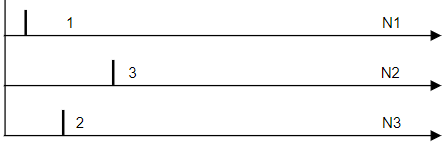}
    \caption{\textcolor{black}{Rank-order coding \newline}}    
\end{subfigure}%
\qquad
\begin{subfigure}{0.5\textwidth}
    \centering
    \includegraphics[scale=0.8]{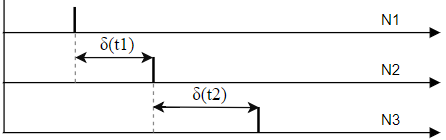}
    \caption{\textcolor{black}{Latency code \newline}}    
\end{subfigure}%

\caption{Different encoding schemes of information in spikes - adapted from \cite{ponulak2011introduction}}
\label{fig:encoding}
\end{figure}

\noindent Generally information available from sensors is not in a form suitable for SNN processing.
Hence, coding such data into spike trains is a major factor in the entire architecture. To address this problem several neural coding strategies based on spike timing have been proposed. Some of these strategies are listed below and visualized in Figure \ref{fig:encoding}. In some cases like event based camera data, data arrives in a form more suitable for SNNs.

1.	\textbf{Time to first spike} – Information is encoded as time between the beginning of stimulus and the time of the first spike in response. As can be seen from Figure 2-a, a group of three neurons N1, N2 and N3 spike in response to a stimulus. The time between the start of the stimulus to the first spike by neuron N2 encodes the type of the stimulus.
Such encoding scheme is generally applied in applications such as artificial tactile and olfactory sensors \cite{chen2011spike}.

2.	\textbf{Rank-order coding (ROC)} – Here, information is coded by the firing order of spikes from the group of neurons that encode the information. As seen in Figure 2-b, the neurons fire in the order N1 followed by N3 and N2 respectively. This sequence of firing of the three neurons encodes the type of stimulus.

3.	\textbf{Latency code} – Information is coded by the difference in time between firing of neurons. It is a highly efficient method of encoding large amounts of information using only a few spikes \cite{borst1999information}. This is because, a slight change in the timings can be used to encode a completely different data sample. Figure 2-c shows the latency between firing of neuron N1 and N2 as 
\begin{math}
    \delta(t1)
\end{math}.
Similarly, the latency in firing of N3 after N2 is depicted as 
\begin{math}
    \delta(t2)
\end{math}. \\

\begin{figure}[!t]
\centering
\begin{subfigure}{0.5\textwidth}
    \centering
    \includegraphics[scale=0.3]{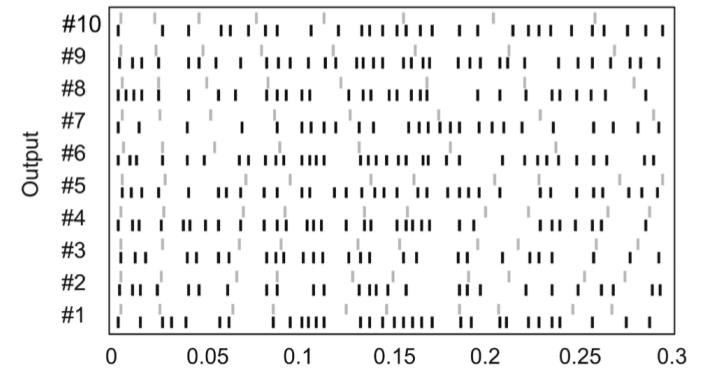}
    \caption{\textcolor{black}{Spike timings before training \newline}}
\end{subfigure}%
\qquad
\begin{subfigure}{0.5\textwidth}
    \centering
    \includegraphics[scale=0.3]{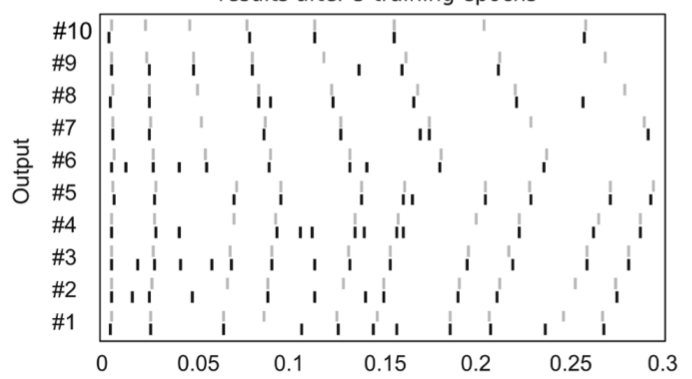}
    \caption{\textcolor{black}{Spike firing rates after 5 epochs training \newline}}    
\end{subfigure}%
\qquad
\begin{subfigure}{0.5\textwidth}
    \centering
    \includegraphics[scale=0.3]{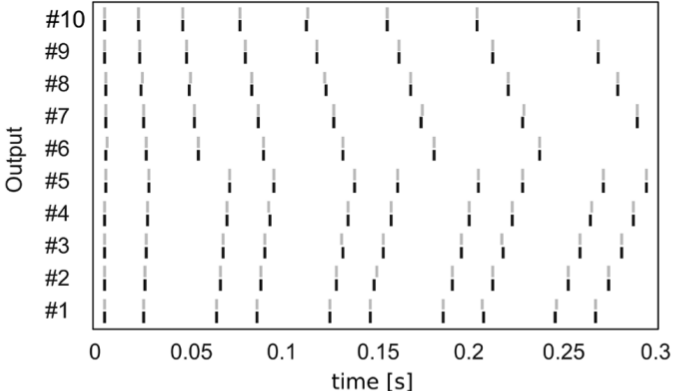}
    \caption{\textcolor{black}{Spike timings after 15 epochs \newline}}    
\end{subfigure}%

\caption{Supervised learning with ReSuMe algorithm. A single-layer feedforward network with 10 spiking neurons. The task is to learn a sample target sequence of spikes assigned individually to each LIF neuron. The gray vertical bars are the target firing times and black bars are the actual time of firing. It can be seen after 15 epochs, the target and actual times are almost identical, describing learning. \cite{ponulak2011introduction}}
\label{fig:resume}
\end{figure}

\noindent  \textbf{Training Spiking Neurons:}
The connections between subsequent neurons are called synapses. In spiking neuron models, these connections or synapses have certain weights or strength associated with them that determine the the strength of the input that the post-synaptic neuron receives from it's pre-synaptic neuron.
These weights can be changed and this phenomenon is called synaptic plasticity.
Several strategies for adjusting the plasticity have been suggested such as depending upon the history of a neuron's response to certain inputs from a particular pre-synaptic neuron.
Other variants may use the simultaneous firing of a pre-synaptic and post-synaptic neuron as a criteria for increasing the synaptic weight etc.
Synaptic plasticity is the key principle by which learning is achieved in SNNs.
Both supervised and unsupervised forms of learning can be modelled using synaptic plasticity.
Figure \ref{fig:resume} shows a sample application of supervised learning using the ReSuMe algorithm. Here the objective was to learn the target firing times of a group of 10 spiking neurons. As seen in the figure, after 15 epochs, most of the neurons have achieved the desired firing times depicted as gray lines. 

We briefly review the main unsupervised learning algorithm. Donald Hebb famously formulated a rule for changing synaptic weights depending on pre-synaptic and post-synaptic activity.
According to Hebb's formula the synaptic weight between neurons i and j, $w_{ji}$, is increased if neurons i and j are simultaneously active.
This method of changing synaptic weights is purely dictated by the input spike train and can lead to pattern recognition in an unsupervised way and no correction based on error evaluation is needed within the network \cite{hinton1999unsupervised,hertz1991introduction}.
The condition formulated by Hebb for increasing or decreasing the synaptic weights between neurons is called  Spike-Timing-Dependent-Plasticity (STDP).



\subsection{CNN vs SNN}


CNNs have shown tremendous progress in their suitability to vision and image based tasks such as image recognition, object detection, pattern recognition.
However, the key elements of the networks, convolution, feature map generation, max pooling etc, involve a lot of matrix multiplication and addition and are compute intensive.
Also, the frame based operation of CNNs involves processing the entire input in a batch, hence individual input channels have to wait till the entire frame of inputs is available. This introduces latency.
Further, the inputs are processed in a layered fashion and an output can only be produced when all layers have finished processing a batch of inputs. This causes latency in the output side.
Due to these latencies and compute intensive operations, inference in data sets such as ImageNet \cite{russakovsky2015imagenet} are not real-time and computationally un-economic.
However, meeting real-time on such targets is mandatory for autonomous driving applications.

Unlike CNNs, SNNs are event based, i.e, events are processed as they are generated. This reduces latency in input processing.
Also, only those input channels are evaluated and processed that have had a change or an event. This reduces the number of inputs that have to be processed in each cycle, as sensors do not typically produce new data on every channel. This reduces computational load and power consumption greatly \cite{farabet2012comparison}.

CNNs can be implemented both in software and in hardware and due to their frame based information processing, the hardware resources can be multiplexed. Thus, higher memory bandwidth and faster data transfer are key for real-time performance.
Unlike CNNs, SNNs process events instead of frames, hence hardware needs to be always available as event generation is not predictive.
Though it may seem to be a limitation, this means, the network is tightly coupled to the hardware and can produce faster response than an equivalent CNN.
To improve the efficiency of a SNN architecture, a modular and re-configurable hardware is more suitable.
\cite{farabet2012comparison}.

\begin{figure}
    \centering
    \includegraphics[scale=0.3]{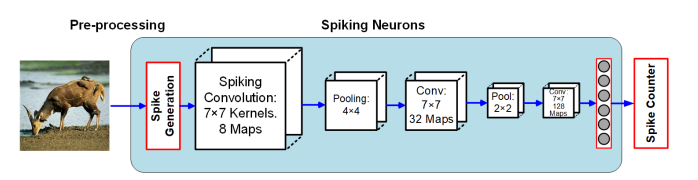}
    \caption{Spiking CNN architecture \cite{cao2015spiking}}
    \label{fig:spikingcnn}
\end{figure}

Given the potential benefits of SNNs, a general question arises on whether CNNs can be adapted to SNNs?
Infact, adapting pre-trained CNNs to equivalent SNNs is easier and produces better results that building a SNN with STDP and unsupervised or supervised learning.
Such adaptations have some key benefits:

1) A spiking convolution operator, analogous to the convolution operator in CNNs would operate much faster due to event based processing, while producing similar results as traditional CNN. 

2) Since events are asynchronous, each convolution operator, supported by its linked modules can operate independent of others, if it has an event for processing. This eliminates the need for a global synchronization among the operators. Such an asynchronous convolution operator may be then implemented as a standard block in hardware for re-usability.

3) Since information is processed on a per-event basis, power is also consumed on a per-event basis. Since sensors typically produce a lot of redundant and sparse data, this could bring a significant reduction in power consumption and computational load.  

Finally SNNs can be queried for results anytime after the first spikes are produced at the output since information processing is not frame based \cite{rueckauer2017conversion}.
Several implementations of deep SNNs on neuromorphic hardware such as SpiNNaker and BrainChip have demonstrated sensor applications that support this potential of SNNs.

Some evidence to support the strong possibilities in research of SNN based networks for object detection is presented in Table \ref{table}.
It is based on an implementation by \cite{rueckauer2017conversion}. It presents a comparison of classification error rates for CNNs and SNN implementation on state of the art data sets \cite{cao2015spiking}.  

\begin{table}[]
\caption{Comparison of ANN and SNN in various computer vision datasets \cite{rueckauer2017conversion}.}
\label{table}
\begin{tabular}{|l|l|l|}
\hline
\textbf{Data set} & \textbf{\begin{tabular}[c]{@{}l@{}}ANN \\ error rate (\%)\end{tabular}} & \textbf{\begin{tabular}[c]{@{}l@{}}SNN \\ error rate (\%)\end{tabular}} \\ \hline
MNIST {[}12{]}    & 0.86                         & 0.86                         \\ \hline
CIFAR-10          & 11.13                        & 11.18                        \\ \hline
ImageNet          & 23.88                        & 25.4                         \\ \hline

\end{tabular}
\end{table}

\section{SNNs in Automated Driving} \label{snn4ad}

\subsection{Use cases in Automated Driving} 

\textbf{Event Driven Computing:} Automated driving has a wide variety of scenarios. At high level, the main scenarios are parking, highway driving and urban driving \cite{heimberger2017computer}. The scene dynamics and understanding is typically different for these scenarios and a customized model is generally used for these scenarios. There are also various scenarios based on weather condition like rainy, day or night, foggy, etc. The combination of various environmental condition is exponential and difficult to have a customized model for each scenario. At the same time, transfer function can be shared across these different scenarios and event triggered mechanism can be used to adapt the regions used. This can be accomplished loosely using shared encoder and gating mechanisms within CNN. However, SNN naturally captures event triggered model. There is a class of cameras called event based cameras which encode information at the sensor level. Recently, deep learning algorithms were demonstrated on event based camera data \cite{maqueda2018event}. \\

\noindent \textbf{Point Cloud:}
Light Detection and Ranging (LiDAR) sensors have recently gained prominence as state of the art sensors in sensing the environment.
They produce a 3D representation of the objects in the field of view as distances of points from the source.
This collection of points over a 3D space is called a 3D Point Cloud.
Though cameras have been used for a long time and they provide a more direct representation of the surrounding, LiDARs have gained ground because of some critical advantages such as long range, robustness to ambient light conditions and accurate localization of objects in 3D space. They produce sparse data and hence suitable for SNNs.



\subsection{Opportunities}

SNNs have shown great potential to either aid or replace CNNs in real-time tasks such as object detection, posture recognition etc.
\cite{hu2016dvs}. Large SNN architectures can be implemented on neuromorphic spiking platforms such as TrueNorth \cite{benjamin2014neurogrid}.
and SpiNNaker \cite{furber2014spinnaker}. The TrueNorth has demonstrated to consume as low as couple hundred mW power while packing a million neurons in it \cite{sawada2016truenorth}.
Driven by the strong motivation to reduce power consumption of integrated circuits, implementations of spiking models have shown to consume in the order of nJ or even pJ \cite{azghadi2014spike} for signal transmission and processing \cite{indiveri2006vlsi}.
Some neuromorphic designs also feature on-chip learning \cite{indiveri2007spike}.

Spiking applications and spike based learning is also suited to dynamic applications like speech recognition systems.
In such systems, training is not sufficient at manufacture as it has to adapt to dynamic conditions such as accents.
Other similar sensors are event based Dynamic Vision Sensor (DVS) \cite{lichtsteiner2008128} \cite{lenero2010signed}.
Some of the applications especially in the object detection and perception based tasks that are of direct relevance to the automotive industry as mentioned briefly below.

1. Object classification on the CIFAR-10 dataset:
\cite{cao2015spiking} designed a Spiking equivalent model of a CNN for object detection on the CIFAR-10 data set.
The CNN was trained on the dataset and the trained model was then converted into spiking with each individual block such as convolution, max pooling, ReLU, being replaced by spiking equivalents.
Their transformed model achieves an error rate of 22.57\%.
CIFAR-10 is a collection of 60,000 labeled images of 10 classes of objects \cite{cao2015spiking} The network architecture is illustrated in Figure \ref{fig:spikingcnn}.

2. Human action recognition:
\cite{zhao2015feedforward} constructed a network to recognize human actions and posture and successfully tested it. The network was trained on an event-based dataset of small video sequences with simple human actions like sitting, walking or bending. They achieved a detection accuracy of 99.48\%.
This work is an indication of how SNNs may be applied to such event based inference tasks. \\

\noindent We summarize the key benefits of SNN for automated driving:

\begin{itemize}
    \item Event driven mechanism which brings adaptation for different scenarios.
    \item Low power consumption when realized as neuromorphic hardware.
    \item Simpler learning algorithm which leads to possibility of on-chip learning for longer term adaptation.
    \item Ability to integrate directly to analog signals leading to tightly integrated system.
    \item Lower latency in algorithm pipeline which is important for high speed braking and maneuvering.
\end{itemize}

\section{Conclusion} \label{conclusion}
Spiking Neural Networks (SNN) are biologically inspired where the neuronal activity is sparse and event driven in order to optimize power consumption. In this paper, we provide an overview of SNN and compare it with CNN and argue how it can be useful in automated driving systems. Overall power consumption  over the driving cycle is a critical constraint which has to be efficiently used especially for electric vehicles. Event driven architectures for various scenarios in automated driving can also have accuracy advantages. 




\bibliographystyle{apalike}
{\small
\bibliography{egbib}

\begin{thebibliography}{}

\bibitem[Azghadi et~al., 2014]{azghadi2014spike}
Azghadi, M.~R., Iannella, N., Al-Sarawi, S.~F., Indiveri, G., and Abbott, D.
  (2014).
\newblock Spike-based synaptic plasticity in silicon: design, implementation,
  application, and challenges.
\newblock {\em Proceedings of the IEEE}, 102(5):717--737.

\bibitem[Benjamin et~al., 2014]{benjamin2014neurogrid}
Benjamin, B.~V., Gao, P., McQuinn, E., Choudhary, S., Chandrasekaran, A.~R.,
  Bussat, J.-M., Alvarez-Icaza, R., Arthur, J.~V., Merolla, P.~A., and Boahen,
  K. (2014).
\newblock Neurogrid: A mixed-analog-digital multichip system for large-scale
  neural simulations.
\newblock {\em Proceedings of the IEEE}, 102(5):699--716.

\bibitem[Bois-Reymond et~al., 1848]{bois1848untersuchungen}
Bois-Reymond, Y. et~al. (1848).
\newblock Investigations on animal electricity {\ "a} t.
\newblock {\em Annalen der Physik}, 151:463--464.

\bibitem[Borst and Theunissen, 1999]{borst1999information}
Borst, A. and Theunissen, F.~E. (1999).
\newblock Information theory and neural coding.
\newblock {\em Nature neuroscience}, 2(11):947.

\bibitem[Cao et~al., 2015]{cao2015spiking}
Cao, Y., Chen, Y., and Khosla, D. (2015).
\newblock Spiking deep convolutional neural networks for energy-efficient
  object recognition.
\newblock {\em International Journal of Computer Vision}, 113(1):54--66.

\bibitem[Cassidy and Ekanayake, 2006]{cassidy2006biologically}
Cassidy, A. and Ekanayake, V. (2006).
\newblock A biologically inspired tactile sensor array utilizing phase-based
  computation.
\newblock In {\em Biomedical Circuits and Systems Conference, 2006. BioCAS
  2006. IEEE}, pages 45--48. IEEE.

\bibitem[Chen et~al., 2011]{chen2011spike}
Chen, H.~T., Ng, K.~T., Bermak, A., Law, M.~K., and Martinez, D. (2011).
\newblock Spike latency coding in biologically inspired microelectronic nose.
\newblock {\em IEEE transactions on biomedical circuits and systems},
  5(2):160--168.

\bibitem[Chou et~al., 2018]{chou2018algorithmic}
Chou, C.-N., Chung, K.-M., and Lu, C.-J. (2018).
\newblock On the algorithmic power of spiking neural networks.
\newblock {\em arXiv preprint arXiv:1803.10375}.

\bibitem[Diehl et~al., 2015]{diehl2015fast}
Diehl, P.~U., Neil, D., Binas, J., Cook, M., Liu, S.-C., and Pfeiffer, M.
  (2015).
\newblock Fast-classifying, high-accuracy spiking deep networks through weight
  and threshold balancing.
\newblock In {\em Neural Networks (IJCNN), 2015 International Joint Conference
  on}, pages 1--8. IEEE.

\bibitem[Escobar et~al., 2009]{escobar2009action}
Escobar, M.-J., Masson, G.~S., Vieville, T., and Kornprobst, P. (2009).
\newblock Action recognition using a bio-inspired feedforward spiking network.
\newblock {\em International Journal of Computer Vision}, 82(3):284.

\bibitem[Farabet et~al., 2012]{farabet2012comparison}
Farabet, C., Paz, R., P{\'e}rez-Carrasco, J., Zamarre{\~n}o, C.,
  Linares-Barranco, A., LeCun, Y., Culurciello, E., Serrano-Gotarredona, T.,
  and Linares-Barranco, B. (2012).
\newblock Comparison between frame-constrained fix-pixel-value and frame-free
  spiking-dynamic-pixel convnets for visual processing.
\newblock {\em Frontiers in neuroscience}, 6:32.

\bibitem[Fu et~al., 2007]{fu2007pattern}
Fu, J., Li, G., Qin, Y., and Freeman, W.~J. (2007).
\newblock A pattern recognition method for electronic noses based on an
  olfactory neural network.
\newblock {\em Sensors and Actuators B: Chemical}, 125(2):489--497.

\bibitem[Furber et~al., 2014]{furber2014spinnaker}
Furber, S.~B., Galluppi, F., Temple, S., and Plana, L.~A. (2014).
\newblock The spinnaker project.
\newblock {\em Proceedings of the IEEE}, 102(5):652--665.

\bibitem[Gerstner and Kistler, 2002]{gerstner2002spiking}
Gerstner, W. and Kistler, W.~M. (2002).
\newblock {\em Spiking neuron models: Single neurons, populations, plasticity}.
\newblock Cambridge university press.

\bibitem[Girshick, 2015]{girshick2015fast}
Girshick, R. (2015).
\newblock Fast r-cnn.
\newblock In {\em Proceedings of the IEEE international conference on computer
  vision}, pages 1440--1448.

\bibitem[Heimberger et~al., 2017]{heimberger2017computer}
Heimberger, M., Horgan, J., Hughes, C., McDonald, J., and Yogamani, S. (2017).
\newblock Computer vision in automated parking systems: Design, implementation
  and challenges.
\newblock {\em Image and Vision Computing}, 68:88--101.

\bibitem[Hertz et~al., 1991]{hertz1991introduction}
Hertz, J., Krogh, A., and Palmer, R.~G. (1991).
\newblock {\em Introduction to the theory of neural computation.}
\newblock Addison-Wesley/Addison Wesley Longman.

\bibitem[Hinton et~al., 1999]{hinton1999unsupervised}
Hinton, G.~E., Sejnowski, T.~J., and Poggio, T.~A. (1999).
\newblock {\em Unsupervised learning: foundations of neural computation}.
\newblock MIT press.

\bibitem[Hu et~al., 2016]{hu2016dvs}
Hu, Y., Liu, H., Pfeiffer, M., and Delbruck, T. (2016).
\newblock Dvs benchmark datasets for object tracking, action recognition, and
  object recognition.
\newblock {\em Frontiers in neuroscience}, 10:405.

\bibitem[Hunsberger and Eliasmith, 2015]{hunsberger2015spiking}
Hunsberger, E. and Eliasmith, C. (2015).
\newblock Spiking deep networks with lif neurons.
\newblock {\em arXiv preprint arXiv:1510.08829}.

\bibitem[Indiveri et~al., 2006]{indiveri2006vlsi}
Indiveri, G., Chicca, E., and Douglas, R.~J. (2006).
\newblock A vlsi array of low-power spiking neurons and bistable synapses with
  spike-timing dependent plasticity.
\newblock {\em IEEE transactions on neural networks}, 17(1).

\bibitem[Indiveri and Fusi, 2007]{indiveri2007spike}
Indiveri, G. and Fusi, S. (2007).
\newblock Spike-based learning in vlsi networks of integrate-and-fire neurons.
\newblock In {\em Circuits and Systems, 2007. ISCAS 2007. IEEE International
  Symposium on}, pages 3371--3374. IEEE.

\bibitem[Lenero-Bardallo et~al., 2010]{lenero2010signed}
Lenero-Bardallo, J.~A., Serrano-Gotarredona, T., and Linares-Barranco, B.
  (2010).
\newblock A signed spatial contrast event spike retina chip.
\newblock In {\em Circuits and Systems (ISCAS), Proceedings of 2010 IEEE
  International Symposium on}, pages 2438--2441. IEEE.

\bibitem[Lichtsteiner et~al., 2008]{lichtsteiner2008128}
Lichtsteiner, P., Posch, C., and Delbruck, T. (2008).
\newblock A 128x128 120 db 15 microsec latency asynchronous temporal contrast
  vision sensor.
\newblock {\em IEEE journal of solid-state circuits}, 43(2):566--576.

\bibitem[Maass, 1997]{maass1997networks}
Maass, W. (1997).
\newblock Networks of spiking neurons: the third generation of neural network
  models.
\newblock {\em Neural networks}, 10(9):1659--1671.

\bibitem[Maqueda et~al., 2018]{maqueda2018event}
Maqueda, A.~I., Loquercio, A., Gallego, G., Garc{\i}a, N., and Scaramuzza, D.
  (2018).
\newblock Event-based vision meets deep learning on steering prediction for
  self-driving cars.
\newblock In {\em Proceedings of the IEEE Conference on Computer Vision and
  Pattern Recognition}, pages 5419--5427.

\bibitem[Ponulak and Kasinski, 2011]{ponulak2011introduction}
Ponulak, F. and Kasinski, A. (2011).
\newblock Introduction to spiking neural networks: Information processing,
  learning and applications.
\newblock {\em Acta neurobiologiae experimentalis}, 71(4):409--433.

\bibitem[Redmon et~al., 2016]{redmon2016you}
Redmon, J., Divvala, S., Girshick, R., and Farhadi, A. (2016).
\newblock You only look once: Unified, real-time object detection.
\newblock In {\em Proceedings of the IEEE conference on computer vision and
  pattern recognition}, pages 779--788.

\bibitem[Rueckauer et~al., 2017]{rueckauer2017conversion}
Rueckauer, B., Lungu, I.-A., Hu, Y., Pfeiffer, M., and Liu, S.-C. (2017).
\newblock Conversion of continuous-valued deep networks to efficient
  event-driven networks for image classification.
\newblock {\em Frontiers in neuroscience}, 11:682.

\bibitem[Russakovsky et~al., 2015]{russakovsky2015imagenet}
Russakovsky, O., Deng, J., Su, H., Krause, J., Satheesh, S., Ma, S., Huang, Z.,
  Karpathy, A., Khosla, A., Bernstein, M., et~al. (2015).
\newblock Imagenet large scale visual recognition challenge.
\newblock {\em International Journal of Computer Vision}, 115(3):211--252.

\bibitem[Sawada et~al., 2016]{sawada2016truenorth}
Sawada, J., Akopyan, F., Cassidy, A.~S., Taba, B., Debole, M.~V., Datta, P.,
  Alvarez-Icaza, R., Amir, A., Arthur, J.~V., Andreopoulos, A., et~al. (2016).
\newblock Truenorth ecosystem for brain-inspired computing: scalable systems,
  software, and applications.
\newblock In {\em Proceedings of the International Conference for High
  Performance Computing, Networking, Storage and Analysis}, page~12. IEEE
  Press.

\bibitem[Sengupta et~al., 2018]{sengupta2018going}
Sengupta, A., Ye, Y., Wang, R., Liu, C., and Roy, K. (2018).
\newblock Going deeper in spiking neural networks: Vgg and residual
  architectures.
\newblock {\em arXiv preprint arXiv:1802.02627}.

\bibitem[Siam et~al., 2017]{siam2017deep}
Siam, M., Elkerdawy, S., Jagersand, M., and Yogamani, S. (2017).
\newblock Deep semantic segmentation for automated driving: Taxonomy, roadmap
  and challenges.
\newblock In {\em Intelligent Transportation Systems (ITSC), 2017 IEEE 20th
  International Conference on}, pages 1--8. IEEE.

\bibitem[Tavanaei et~al., 2018]{tavanaei2018deep}
Tavanaei, A., Ghodrati, M., Kheradpisheh, S.~R., Masquelier, T., and Maida,
  A.~S. (2018).
\newblock Deep learning in spiking neural networks.
\newblock {\em arXiv preprint arXiv:1804.08150}.

\bibitem[Thorpe et~al., 2001]{thorpe2001spike}
Thorpe, S., Delorme, A., and Van~Rullen, R. (2001).
\newblock Spike-based strategies for rapid processing.
\newblock {\em Neural networks}, 14(6-7):715--725.

\bibitem[Wunderlich et~al., 2018]{wunderlich2018demonstrating}
Wunderlich, T., Kungl, A.~F., Hartel, A., Stradmann, Y., Aamir, S.~A.,
  Gr{\"u}bl, A., Heimbrecht, A., Schreiber, K., St{\"o}ckel, D., Pehle, C.,
  et~al. (2018).
\newblock Demonstrating advantages of neuromorphic computation: A pilot study.
\newblock {\em arXiv preprint arXiv:1811.03618}.

\bibitem[Zhao et~al., 2015]{zhao2015feedforward}
Zhao, B., Ding, R., Chen, S., Linares-Barranco, B., and Tang, H. (2015).
\newblock Feedforward categorization on aer motion events using cortex-like
  features in a spiking neural network.
\newblock {\em IEEE Trans. Neural Netw. Learning Syst.}, 26(9):1963--1978.

\bibitem[Zhou and Wang, 2018]{zhou2018object}
Zhou, S. and Wang, W. (2018).
\newblock Object detection based on lidar temporal pulses using spiking neural
  networks.
\newblock {\em arXiv preprint arXiv:1810.12436}.

\end{thebibliography}
}

\end{document}